\documentclass{article}
\usepackage{spconf,amsmath,graphicx}
\usepackage{amsmath,amssymb,amsfonts}
\usepackage{pifont}
\usepackage[colorlinks=true,citecolor=blue]{hyperref}
\usepackage{url}
\usepackage{subcaption}
\usepackage[font=small]{caption}
\usepackage{bm}
\usepackage{tabularx}
\usepackage[table]{xcolor}
\usepackage{capt-of}

\newcolumntype{Y}{>{\centering\arraybackslash}X}

\newcommand{\rk}[1]{{\color{black}{#1}}}
\newcommand{\xy}[1]{{\color{black}{#1}}}

\title{FAST PREDICTIVE MULTIMODAL IMAGE REGISTRATION}
\name{Xiao Yang$^1$, Roland Kwitt$^4$, Martin Styner$^{1,3}$, Marc Niethammer$^{1,2}$}
\address{$^1$Department of Computer Science, UNC Chapel Hill, USA\\
  $^2$Biomedical Research Imaging Center, UNC Chapel Hill, USA\\
  $^3$Department of Psychiatry, UNC Chapel Hill, USA\\
  $^4$Department of Computer Science, University of Salzburg, Austria}
\begin{document}
%\ninept
%
\maketitle
\begin{abstract}
We introduce a \rk{deep encoder-decoder architecture} for image deformation prediction  from \emph{multimodal images}. Specifically, we design an image-patch-based 
deep network that jointly (i) learns an image similarity measure and (ii) the relationship between image patches and deformation parameters. While our method can be applied to general image registration formulations, we focus on the Large Deformation Diffeomorphic Metric Mapping (LDDMM) registration model. By predicting the \emph{initial momentum} of the shooting formulation of LDDMM, we preserve its mathematical properties and drastically reduce the computation time, compared to optimization-based approaches. Furthermore, we create a Bayesian probabilistic version of the network that allows evaluation of registration uncertainty via sampling of the network at test time. We evaluate our method on a 3D brain MRI dataset using both T1- and T2-weighted images. Our experiments show that our method generates accurate predictions and that {\it learning} the similarity measure leads to more consistent registrations than relying on {\it generic} multimodal image similarity measures, such as mutual information. Our approach is an order of magnitude faster than optimization-based LDDMM.
\end{abstract}
\section{INTRODUCTION}
Multimodal image registration seeks to estimate spatial correspondences between image pairs from different imaging modalities (or protocols). In general image registration, these correspondences are estimated by finding the spatial transformation which makes a transformed source image most {\it similar} to a fixed target image. For unimodal image registration, image similarity should be high if images are close to identical, which allows using simple image similarity measures such as the sum of squared intensity differences (SSD) between image pairs. Assessing image similarity {\it across} modalities is substantially more difficult as image appearance can vary widely, e.g., due to different underlying physical imaging principles. In fact, these differences are {\it desired} as they can, for example, highlight different tissue properties in brain imaging. Hence, more sophisticated multimodal similarity measures are required. Furthermore, image registration results are driven by both the chosen similarity measure {\it and} the chosen deformation model. Hence, especially for multimodal image registration, where assessing image similarity becomes challenging, considering the similarity measure {\it jointly} with the deformation model is important.\\
\indent The most popular multimodal similarity measure is mutual information (MI)~\cite{viola1997}, but other {\it hand-crafted} multimodal similarity measures have also been proposed~\cite{Meyer1998, Hermosillo2002, Lorenzen2006}. These approaches {\it assume} properties characterizing good image alignment, but do not {\it learn} them from data. Hence, more recently, learning-based approaches to measure image similarity have been proposed. These techniques include measuring image similarity by comparing observed and learned intensity distributions via KL-divergence~\cite{Guetter2005}; or learning the similarity of image pixels/voxels or image features (e.g., Fourier/Gabor features) via max-margin structured output learning~\cite{Lee2009}, boosting~\cite{Michel2011}, or deep learning~\cite{Cheng2015,simonovsky2016}. \xy{Some methods avoid a complex similarity measure by applying image synthesis for the source/target image to change the task to unimodal registration~\cite{roy2013magnetic,jog2013magnetic,VanNguyen2015}. However, the registration performance then heavily depends on the synthesis accuracy.}\\ 
\indent Once a similarity measure (learned or hand-crafted) is chosen, registration 
still requires finding the optimal registration parameters by numerical optimization. This can be particularly costly for popular nonparametric elastic or fluid registration approaches which are based on models from continuum mechanics~\cite{holden2008,modersitzki2004} and therefore require the optimization over functions, which results in millions of unknowns after numerical discretization. Approaches to avoid numerical optimization by prediction have been proposed. But, unlike our proposed method, they either focus on predicting displacements via optical flow~\cite{deepflow, flownet} or low-dimensional parametric models~\cite{chou20132d,Wang201561,Becker16}, or cannot address multimodal image registration~\cite{yang2016}, or do not consider jointly learning a model for deformation prediction and image similarity~\cite{Tian2015}. \\
\indent Specifically, Chou et al.~\cite{chou20132d} propose a multi-scale linear regressor for affine transformations or low-dimensional parameterized deformations via principal component analysis. Wang et al.~\cite{Wang201561} introduce a framework that involves key-point matching using sparse learning, followed by dense deformation field interpolation, which heavily depends on the accuracy of key-point selection. Cao et al.~\cite{Tian2015} propose a semi-coupled dictionary learning approach to jointly model unimodal image appearance and deformation parameters, but only a linear relationship between the two is assumed. The two closest methods to our proposed approach are~\cite{yang2016,Becker16}. Yang et al. \cite{yang2016} use deep learning to model the nonlinear relationship between image appearance and LDDMM deformation parameters, but only consider unimodal atlas-to-image registration, instead of general image-to-image registration. Guti\'erez-Becker et al. \cite{Becker16} learn a multimodal similarity measure using a regression forest with Haar-like features combined with a prediction model for a low-dimensional parametric  B-spline model (5 nodes/dimension). In contrast, our approach (i) predicts the initial momentum of the shooting formulation of LDDMM~\cite{Vialard2012}, a nonparametric registration model\footnote{The relaxation formulation of LDDMM~\cite{beg2005} is for example the basis of the successful ANTs~\cite{avants2011} registration tools.}; and (ii) jointly learns a multimodal similarity measure from image-patches without requiring feature selection.

\noindent
\textbf{Contributions.} We propose a deep learning architecture to \emph{jointly} learn a multimodal similarity measure and the relationship between images and deformation parameters. Specifically, we design a deep encoder-decoder network to predict the initial momentum of LDDMM using multimodal image patches for \emph{image-to-image registration} (opposed to atlas-to-image registrations as in~\cite{yang2016}). We 
focus on LDDMM, but our method is applicable to other registration models. Our contributions are: (i) a patch-based deep network that generates accurate deformation parameter predictions using multimodal images; (ii) the simultaneous learning of a multimodal image similarity measure based only on multimodal image patches; (iii) an order of magnitude speedup, compared to optimization-based (GPU-accelerated) registration of 3D images; and (iv) a Bayesian extension of our model to provide uncertainty estimates for predicted deformations.

\noindent
\textbf{Organization.} Sec.~\ref{sec:LDDMM} reviews the initial momentum formulation of LDDMM and discusses our motivation for this parameterization. Sec.~\ref{sec:network} introduces our deterministic and Bayesian network structure, as well as our method of speeding up the computation. Sec.~\ref{sec:experiments} presents experimental results on a 3D autism brain dataset. Sec.~\ref{sec:discussion} discusses potential future research directions, experiments, and possible extensions.

\section{INITIAL MOMENTUM LDDMM SHOOTING}
\label{sec:LDDMM}
Given a moving image $S$ and a target image $T$, LDDMM estimates a diffeomorphism $\varphi$ such that $S\ \circ\ \varphi^{-1} \approx T$, where $\varphi\ \dot{=}\ \phi(1)$ is generated via a smooth flow $\phi(t), t\in[0,1]$. The parameter for the LDDMM shooting formulation is the \emph{initial momentum vector field} $m(t)$, which is used to compute $\phi(t)$. The initial momentum is the dual of the initial velocity field $v(t)$ which is an element of a reproducing kernel Hilbert space $V$. The vector fields $m$ and $v$ are connected by a positive definite self-adjoint smoothing kernel $K$ via $v = Km$ and $m=Lv$, where $L$ denotes the inverse of $K$. The energy function for the shooting formulation of LDDMM is~\cite{singh2013, Vialard2012}
\begin{equation}
    E(m_0) = \langle m_0, Km_0\rangle + \frac{1}{\sigma^2}||S\circ \phi^{-1}(1)- T||^2, \text{where}
    \label{eqn:momentum_energy}
\end{equation}
\begin{equation}
\begin{split}
m_t + \text{ad}_v^*m = 0, m(0) = m_0, \\ \phi^{-1}_t + D \phi^{-1} v = 0,\quad  \phi^{-1}(0)=\text{id}, \quad m - Lv = 0.
\label{eqn:forward}
\end{split}
\end{equation}
Here, $\text{id}$ is the identity map, $\text{ad*}$ is the negated Jacobi-Lie bracket of vector fields, i.e., $\text{ad}_v m = Dmv - Dvm$,  $D$ denotes the Jacobian operator, and subscript $t$ denotes the derivative w.r.t. time $t$. As in~\cite{yang2016}, we choose to \emph{predict the initial momentum $m_0$}. This is motivated by the observation that the momentum parameterization (unlike parameterization via displacement or vector fields) allows patch-wise prediction, because the momentum is non-zero only on image edges (in theory, $m=\lambda\nabla I$ for images, where $\lambda$ is a scalar field). Thus, from a theoretical point of view, no information is needed outside the patch for momentum prediction, and $m=0$ in homogeneous regions. Furthermore, deformation smoothness is guaranteed via $K$ \emph{after} the prediction step. I.e., given a sufficiently strong regularizer, $L$, diffeomorphic deformations are obtained by integrating Eq.~\eqref{eqn:forward}.

\section{NETWORK STRUCTURE}
\label{sec:network}
\begin{figure}[tb]
\centering
\includegraphics[width=.7\columnwidth]{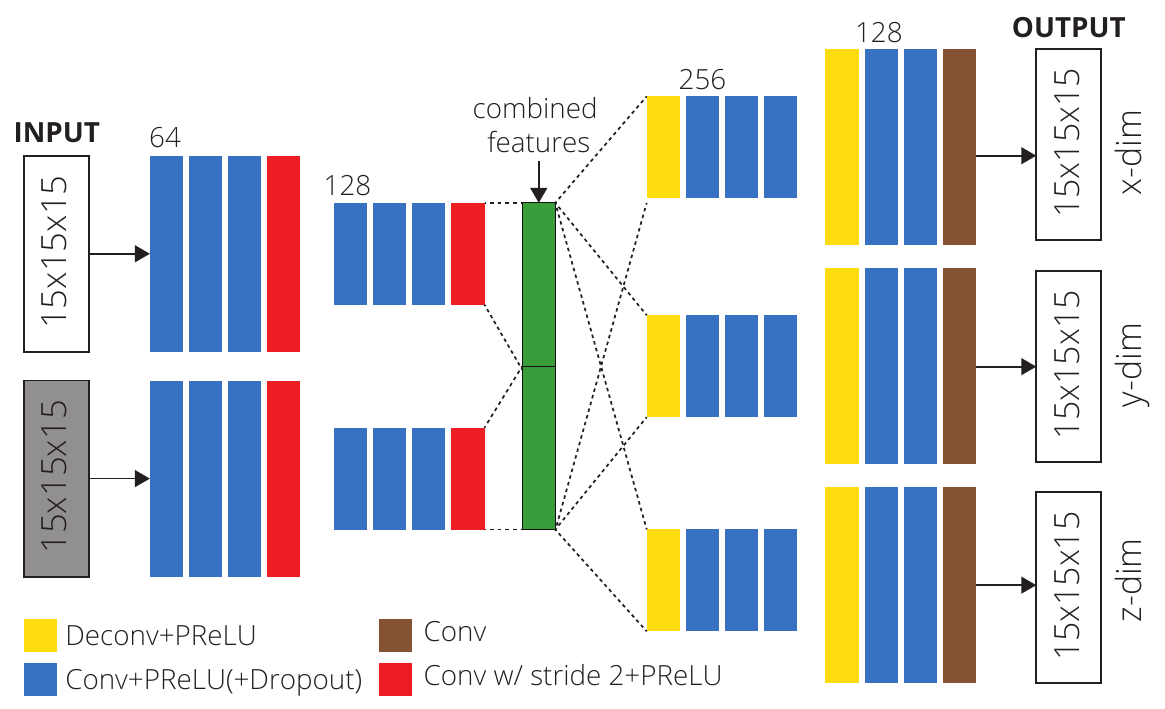}
\caption{3D Bayesian network structure. The network takes 2 3D patches from moving/target images and generates 3 3D initial momentum patches (in 3 dim.). Numbers for the input and output denote input/output patch size, and numbers in the network indicate the number of features (output channels) for the conv. layers. To create a deterministic network, we remove all dropout layers.}
\label{fig:network}
\vspace{-0.4cm}
\end{figure}

Fig.~\ref{fig:network} shows our network structure for 3D multimodal image deformation prediction. This network is an encoder-decoder network, where \emph{two} encoders compute features from the moving/target image patches independently. The learned features from the two encoders are simply concatenated and sent to \emph{three} decoders to generate one initial momentum patch for each dimension. \xy{The two encoder structure is different from the network in \cite{yang2016}} since using two encoders instead of a single encoder with two initial input channels 
%and double feature space 
has the effect of reducing overfitting, as shown in Sec.~\ref{sec:experiments}. All convolutional layers use $3\times3\times3$ filters, and we choose PReLU~\cite{PReLU} as the non-linear activation layer. Pooling and unpooling layers are usually used for multi-scale image processing, \xy{but unlike~\cite{yang2016} where unpooling based on pooling index is straightforward}, they are not suitable for our formulation since the two encoders perform pooling independently. Therefore, we follow the idea of~\cite{SpringenbergDBR14} and use convolutional layers with a stride of 2 and deconvolution layers~\cite{Long2015} as surrogates for pooling and unpooling layers. I.e., the network learns its pooling and unpooling operations during training. For training, we use the L1 norm as our similarity criterion. To predict the full image momentum during testing, we use a sliding window approach, predicting the momentum patch-by-patch, followed by averaging the momenta in overlapping areas. We refer to this architecture as our \emph{deterministic network.}

%The network architecture discussed above is for the deterministic version of our network. 
We extend the network to a \emph{Bayesian network} by adding dropout layers~\cite{srivastava14a} after all convolution+PReLU layers except for the ``pooling/unpooling" (convolutional) layers~\cite{Gal2015Bayesian}. Given the input($I$)/output($O$) of a convolutional layer and the convolutional layer's weight, $W$, adding dropout can be regarded as performing variational inference for the posterior distribution of the conv layer's weight $p(W|I, O)$ using a Bernoulli distribution $q(W) = W * (\text{Bernoulli}(p)_i)|_{i=1}^N$, where $p$ is the dropout probability, $N$ is the number of nodes in the conv layer, and $\text{Bernoulli}(p)_i$ randomly sets the $i$th node in the conv layer to 0. According to \cite{Gal2015Bayesian}, training this dropout network is equivalent to minimizing the KL-divergence between the variational and true posterior $\text{KL}(q(W)|p(W|I, O))$. During testing, we keep the dropout layers and sample the network to obtain multiple momentum predictions for a single image. We choose the sample mean as the final momentum prediction and perform LDDMM shooting (by integrating Eq.~\eqref{eqn:forward}) for the momentum samples to generate multiple deformation fields. The variances of the deformation fields then estimate the predicted deformation \emph{uncertainty}. We set the dropout probability to $0.3$.

\noindent
\textbf{Patch pruning.}
We achieve a substantial speedup in computation time by using a \emph{patch pruning} strategy. Specifically, we apply a large sliding window stride and ignore patches only containing the background of both the moving image and the target image. In our experiments, these technique reduced the number of patches to predict by approximately 99.85\% for $229\times193\times193$ 3D brain images with $15\times15\times15$ patch size and a sliding window stride of 14.

\vspace{-0.2cm}
\section{EXPERIMENTS}
\label{sec:experiments}
%\noindent\begin{minipage}{\linewidth}
\begin{figure*}[t!]
%\begin{table*}[t]
  \footnotesize
\centering
\begin{tabularx}{0.78\textwidth}{|r|c *{6}{|Y}|}
\hline
& \multicolumn{7}{c|}{\textbf{Deformation Error w.r.t LDDMM optimization on T1w-T1w data} [voxel]}\\ \hline
\textsl{Data Percentile} & 0.3\% & 5\% & 25\% & 50\% & 75\% & 95\% & 99.7\%\\ \hline
Affine (Baseline) & 0.1664 & 0.46 & 0.9376 & 1.4329 & 2.0952 & 3.5037 & 6.2576\\ \hline
\textbf{Ours}, T1w-T1w data  & \cellcolor{green!40}{0.0353} & \cellcolor{green!40}{0.0951} & \cellcolor{green!40}{0.1881} & \cellcolor{green!40}{0.2839} & \cellcolor{green!40}{0.416} & \cellcolor{green!40}{0.714} & \cellcolor{green!40}{1.4409} \\ \hline
\cite{yang2016}, T1w-T2w data & 0.0582 & 0.1568 & 0.3096 & 0.4651 & 0.6737 & 1.1106 & 2.0628 \\ \hline
\textbf{Ours}, T1w-T2w data & \cellcolor{green!40}{0.0551} & \cellcolor{green!40}{0.1484} & \cellcolor{green!40}{0.2915} & \cellcolor{green!40}{0.4345} & \cellcolor{green!40}{0.6243} & \cellcolor{green!40}{1.0302} & \cellcolor{green!40}{2.0177} \\ \hline
\textbf{Ours}, T1w-T2w data, 10 images & \cellcolor{green!40}{0.0663} & \cellcolor{green!40}{0.1782} & \cellcolor{green!40}{0.3489} & \cellcolor{green!40}{0.5208} & \cellcolor{green!40}{0.752} & \cellcolor{green!40}{1.2421} & \cellcolor{green!40}{2.3454} \\ \hline
& \multicolumn{7}{c|}{\textbf{Prediction/Optimization error between T1w-T2w and T1w-T1w} [voxel]} \\ \hline
\textsl{Data Percentile} & 0.3\% & 5\% & 25\% & 50\% & 75\% & 95\% & 99.7\% \\ \hline
\textbf{Ours} & \cellcolor{green!40}{0.0424} & \cellcolor{green!40}{0.1152}  & \cellcolor{green!40}{0.2292} & \cellcolor{green!40}{0.3444} & \cellcolor{green!40}{0.4978} & \cellcolor{green!40}{0.8277} & \cellcolor{green!40}{1.6959}\\ \hline
\texttt{NiftyReg} (Baseline) & 0.2497 & 0.7463 & 1.8234 & 3.1719 & 5.1124 & 8.9522 & 14.4666 \\ \hline
\end{tabularx}
\captionof{table}{Evaluation results for the 3D dataset.}
\label{table:3D}
%\end{table*}
%\begin{figure*}[th]
\centering
\includegraphics[width=0.78\textwidth]{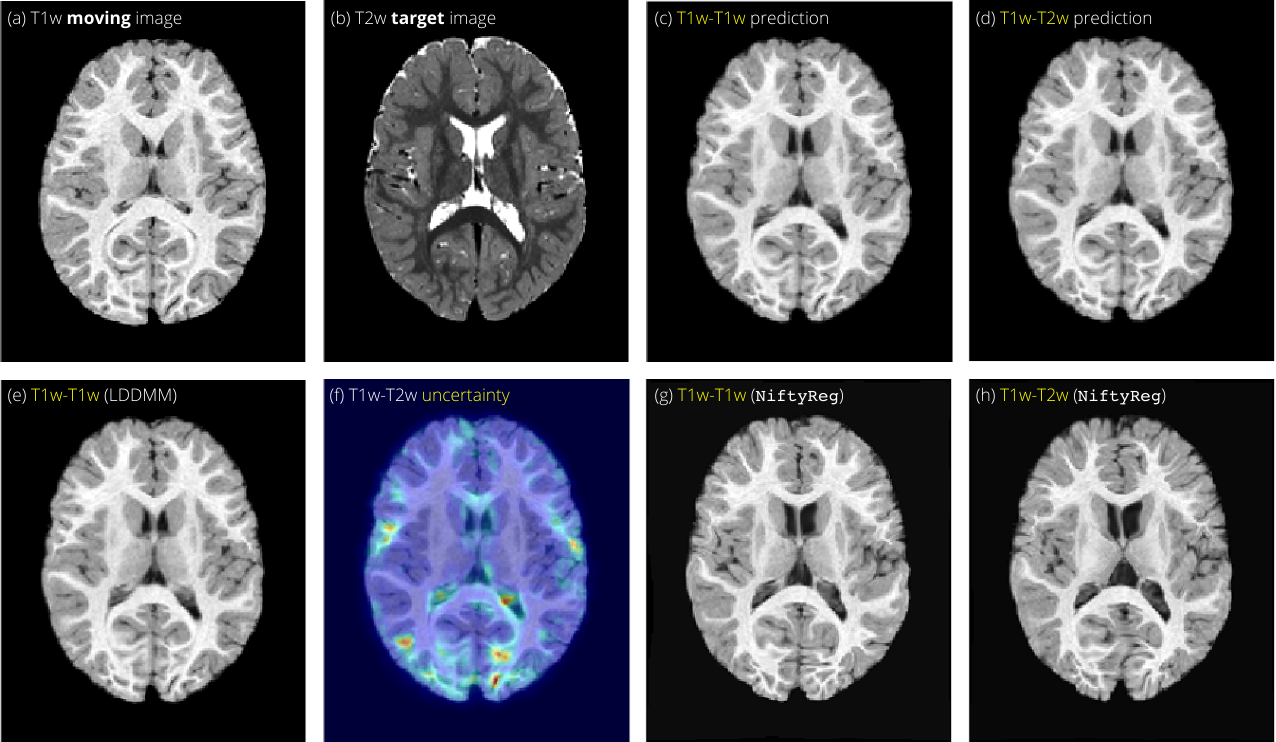}
\captionof{figure}{Exemplary test case. (a) T1w moving image; (b) T2w target image; (c)-(d) deformation prediction result from T1w-T1w/T1w-T2w data; (e) deformation result by LDDMM optimization for T1w-T1w registration; (f) uncertainty of predicted T1w-T2w deformation as the 2-norm of the sum of variances of deformation fields in all spatial directions, mapped on the predicted T1w-T2w wrapped image. Yellow = more uncertainty, blue = less uncertainty; (g)+(h) \texttt{NiftyReg} registration result for T1w-T1w/T1w-T2w pair.}
\label{fig:example}
\vspace{-0.3cm}
%\end{figure*}
\end{figure*}
%\end{minipage}
We assess our approach on the IBIS 3D Autism Brain image dataset~\cite{Wolff2046}, 
containing 375 T1w/T1w brain images ($229\times193\times193$) of 2 years subject with pediatric disease. We first register all images affinely to the ICBM 152 atlas~\cite{FonovEBAMC11} and select 359 images for training and the remaining 16 images for testing. For training, we randomly select 359 T1w-T1w image pairs and perform LDDMM registration using PyCA\footnote{\url{https://bitbucket.org/scicompanat/pyca}} on GPU with SSD as image similarity measure. We set the parameters for the LDDMM regularizer $L = a\Delta^2 + b\Delta + c$ to $[a, b, c] = [0.01, 0.01, 0.001]$, and $\sigma$ in \xy{Eqn.\ref{eqn:momentum_energy}} to 0.2. \emph{We then train the network to predict the momenta generated from the T1w-T1w registrations using their corresponding T1w and T2w images}. The network is implemented in \texttt{Torch} on a TITAN X GPU and is optimized (over 10 epochs) using \texttt{rmsprop} with a learning rate of $= 0.0001$, momentum decay $=0.1$ and update decay $=0.01$. For testing, we perform T1w-T2w pairwise registration predictions for all 16 test images, excluding self-registrations. This results in a total of 240 test cases. Each prediction result is compared to the T1w-T1w registrations obtained via LDDMM optimization (used as ground truth). 
The patch size for the 3D network is $15\times15\times15$ and we use a sliding window size of $14$ for both training and testing. For comparison to the ground truth deformation from LDDMM optimization, we trained another network using T1w-T1w data to perform prediction on the T1w-T1w registration cases. This network serves as the ``upper limit'' of our multimodal network's potential performance. We also implemented the architecture from~\cite{yang2016} and train the network using the T1w-T2w data for comparison. The deformation errors are calculated as the 2-norm of the voxel-wise difference between the predicted deformations and the deformations obtained from LDDMM optimization.\\
\indent Tab.~\ref{table:3D}(top) lists the evaluation results: our multimodal network (T1w-T2w) greatly reduces deformation error compared to affine registration, and only has a slight accuracy loss compared to the T1w-T1w network. This demonstrates registration consistency (to the T1w-T1w registration result) of our approach achieved by {\it learning} the similarity measure between the two modalities. Moreover, the deformation error data percentiles of Tab.~\ref{table:3D} show that our network achieves a slight deformation error decrease of $2.1\%{\sim} 7.3\%$ for all data percentiles compared to \cite{yang2016}. This is likely due to less overfitting by using two small encoders.\\
\indent We also test our network for registration tasks with limited training data. To do so, we randomly choose only 10 out of the 359 training images to perform pairwise registration, generating 90 T1w-T1w registration pairs. We then use these 90 registrations to train our T1w-T2w network model. %The result is shown as the `\textbf{Ours}, T1w-T2w data, 10 image' row in Table~\ref{table:3D}.
Tab.~\ref{table:3D} shows that although the network used only 10 images for training, performance only decreases slightly in comparison to our T1w-T2w network using 359 image pairs for training. Hence, by using patches, our network model can also be successfully trained with limited training images.\\
\indent To further test our network's consistency in relation to the T1w-T1w prediction results, we calculate the deformation error of our T1w-T2w network w.r.t the T1w-T1w network. For comparison, we also run \texttt{NiftyReg}~\cite{Modat2010278} B-spline registration on both T1w-T1w and T1w-T2w test cases using normalized mutual information (NMI) with a grid size of 4 and a bending energy weight of 0.0001; we compare the deformation error between T1w-T2w and T1w-T1w registrations, see Tab.~\ref{table:3D}(bottom). Compared to \texttt{NiftyReg}, our method is more consistent for multimodal prediction. Fig.~\ref{fig:example} shows one test case: using \texttt{NiftyReg} generates large differences in the ventricle area between the T1w-T1w and T1w-T2w cases, while our approach does not. We attribute this result to the shortcomings of NMI and not to \texttt{NiftyReg} as a registration method. We also computed the 2-norm of the sum of variances of deformation fields in all directions as the uncertainty of the deformation, shown in Fig.~\ref{fig:example}(f). We observe high uncertainty around the ventricle, due to the drastic appearance change in this area between the moving and the target image.\\
\noindent\textbf{Computation time}. On average, our method requires 24.46s per case. Compared to (GPU) LDDMM optimization, we achieve a 36x speedup. Further speedups can be achieved by using multiple GPUs for independent patch predictions.
\vspace{-0.2cm}
\section{DISCUSSION AND SUPPORT}
\label{sec:discussion}
We proposed a fast method for multimodal image registration which simultaneously (i) learns the multimodal image similarity measure from image-patches and (ii) predicts registrations based on LDDMM, thereby guaranteeing diffeomorphic transformations. \xy{Different from~\cite{yang2016}, we use a different network structure for multimodal instead of unimodal image registration, and choose a new strategy to train the network.} Our method shows good prediction performance and high consistency of the multimodal registration result in comparison to unimodal registration. Future work should test the registration performance via landmarks or volumetric overlap measures. Comparisons to other registration approaches and a direct LDDMM implementation with a standard multimodal similarity measure such as MI would also be desirable.\\
\indent This work is supported by NSF EECS-1148870, NIH 1 R41 NS091792-01, R01-HD055741, R01-HD059854 and U54-HD079124. We thank NVIDIA for the GPU donation.

%Our method could be directly applied to non-parametric registration methods with voxel-wise parameters, and experiments on such methods (e.g., diffeomorphic demons, stationary LDDMM) will be conducted in the future. Besides, additional experiments comparing our method with a LDDMM implementation with a powerful multimodal similarity measure would be useful to further evaluate robustness.
%of our framework.

\let\oldbibliography\thebibliography
\renewcommand{\thebibliography}[1]{%
  \oldbibliography{#1}%
  \setlength{\itemsep}{0pt}%
}

\vspace{-0.2cm}

\small
\bibliographystyle{IEEEbib}
\bibliography{strings,refs}

\end{document}